\pdfoutput=1

\documentclass[11pt]{article}

\usepackage[preprint]{acl}

\usepackage{times}
\usepackage{latexsym}
\usepackage{array,booktabs,multirow,xcolor}
\usepackage{amssymb}
\usepackage{amsmath}
\usepackage{tabularx}
\usepackage{enumitem}
\usepackage{amsfonts}
\usepackage{bbm}

\usepackage[T1]{fontenc}

\usepackage[utf8]{inputenc}

\usepackage{microtype}

\usepackage{inconsolata}

\usepackage{graphicx}
\usepackage{amsmath}
\usepackage{booktabs}

\title{When to Trust Context: Self-Reflective Debates for Context Reliability}

\author{
 \textbf{Zeqi Zhou\textsuperscript{1,*}},
 \textbf{Fang Wu\textsuperscript{2,*,$\dagger$}},
 \textbf{Shayan Talaei\textsuperscript{2,*}},
 \textbf{Haokai Zhao\textsuperscript{3}},\\
 \textbf{Cheng Meixin\textsuperscript{4}},
 \textbf{Tinson Xu\textsuperscript{5}},
 \textbf{Amin Saberi\textsuperscript{2}},
 \textbf{Yejin Choi\textsuperscript{2}}\\
 \textsuperscript{1}Brown University,
 \textsuperscript{2}Stanford University,
 \textsuperscript{3}University of New South Wales,\\
 \textsuperscript{4}Xi'an University of Electronic Science and Technology,
 \textsuperscript{5}University of Chicago\\
 \small{
   *Equal contribution. \quad
   \textbf{Correspondence:} \href{mailto:fangwu97@stanford.edu}{fangwu97@stanford.edu}
 }
}

\begin{document}
\maketitle
\begin{abstract}
    Large language models frequently encounter conflicts between their parametric knowledge and contextual input, often resulting in factual inconsistencies or hallucinations. We propose Self-Reflective Debate for Contextual Reliability (SR-DCR), a lightweight framework that integrates token-level self-confidence with an asymmetric multi-agent debate to adjudicate such conflicts. A critic, deprived of context, challenges a defender who argues from the given passage; a judge model evaluates the debate and determines the context's reliability. The final answer is selected by combining the verdict with model confidence. Experiments on the ClashEval benchmark demonstrate that SR-DCR consistently enhances robustness to misleading context while maintaining accuracy on trustworthy inputs, outperforming both classical debate and confidence-only baselines with minimal computational overhead. The code is available at~\url{https://github.com/smiles724/Self-Reflective-Debates}. 
\end{abstract}
\begin{figure*}[t]
  \centering
  \includegraphics[width=\textwidth]{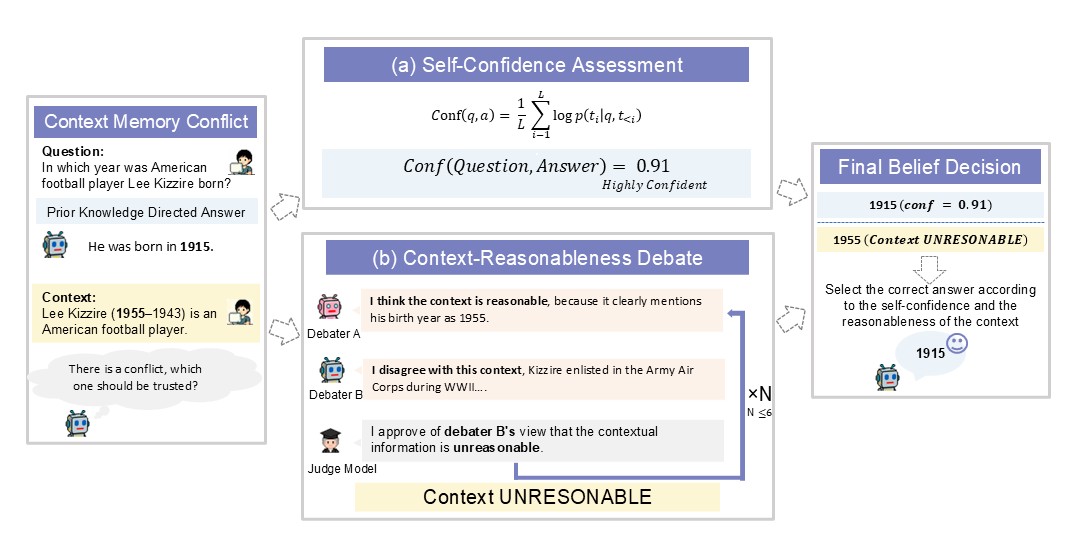}
  \caption{Overview of \textbf{SR-DCR} (Self-Reflective Debate for Contextual Reliability), a debate-driven framework for resolving conflicts between parametric priors and contextual evidence. When the model’s prior belief conflicts with the external passage (left), SR-DCR evaluates two key signals: (a) the self-confidence of the zero-context answer \(\hat{a}_{\textsc{prior}}\); and (b) the context's reasonableness, determined via an asymmetric debate between two LLM agents. A judge model monitors the debate over \(N\) rounds and issues a verdict on the trustworthiness of the context. The final answer is selected by jointly considering both the model's self-confidence and the judged context reliability (right).}
  \label{fig:SR-DCR method}
  \vspace{-1em}
\end{figure*}
\section{Introduction}

Large language models (LLMs)~\citep{wang2024survey,guo2024large} reason using two distinct sources of knowledge: \textit{parametric knowledge}, stored in model weights, and \textit{contextual input}, provided at inference time. This dual-source capability grants LLMs remarkable flexibility, but also exposes a critical vulnerability: when internal priors and external context conflict, models frequently generate hallucinated or factually inconsistent outputs. These failures range from parroting erroneous passages (e.g., “people eat rocks”) to relying on outdated beliefs (e.g., mislocating the Eiffel Tower in Rome). 

Such errors highlight the absence of a principled mechanism for reconciling contradictions between knowledge sources. Empirical studies show that LLMs often defer to memorized facts even when these subtly contradict the context~\citep{brown2020language,longpre2021entity}. Retrieval-augmented generation (RAG) systems compound the problem by amplifying misleading context when it appears superficially fluent or plausible~\citep{niu2023ragtruth}. Reliance on either source in isolation leads to brittle and unpredictable behavior.

Recent work has explored token-level confidence as a signal for epistemic reliability. For example, \citet{wu2024clasheval} compare token probabilities with and without context to select the more trustworthy response. More broadly, output confidence---estimated from log-probabilities---correlates with familiarity: confident predictions tend to reflect well-known facts, while uncertain answers often correspond to ambiguous or unfamiliar queries~\citep{wang2022self}.

In parallel, \textit{multi-agent debate} (MAD)~\citep{du2023improving,chan2023chateval,liang2023encouraging,kenton2024scalable,lang2025debate,agarwal2025when} has emerged as a promising oversight mechanism. By prompting agents to critique each other, it enables deliberation, mitigates hallucinations, and often allows even weak judges to identify correct answers~\citep{irving2018ai,michael2023debate}. However, existing MAD frameworks typically assume consistent inputs across agents, and rarely examine how debates unfold when internal priors and external context offer conflicting evidence.

\vspace{0.5em}
\noindent\textbf{Our Contributions.}  
This work addresses the open challenge of resolving internal–external knowledge conflict in LLMs. We contribute:
\begin{enumerate}[leftmargin=*, itemsep=2pt]
    \item A systematic analysis of model behavior under graded contextual perturbations, revealing how token-level confidence and factual familiarity shape LLM preferences for internal versus external knowledge.
    \item An evaluation of standard MAD setups under adversarial conflict, showing that symmetric debate alone often fails to arbitrate between incompatible sources.
    \item A novel framework, \textbf{Self-Reflective Debate for Contextual Reliability (SR-DCR)}, illustrated in Fig.~\ref{fig:SR-DCR method}. SR-DCR combines token-level self-confidence with an \textbf{Asymmetric Context Verification Debate (ACVD)}, in which one agent defends the context and another, deprived of it, argues from prior knowledge. A judge resolves the debate over multiple rounds, and a final decision rule integrates both context reliability and confidence to select the answer or abstain. This framework improves robustness and factual accuracy across multiple QA benchmarks.
\end{enumerate}

\section{Preliminaries}
\subsection{Background of Knowledge Conflict}
\paragraph{Problem Statement.} We investigate how LLMs reconcile conflicting signals between their internal factual priors and externally provided context. Specifically, each instance is defined as a triple \((q, a, c)\), where:
\begin{itemize}[itemsep=0pt, topsep=0pt, leftmargin=*, labelsep=0.5em]
    \item \(q\) is a question generated from a factual \emph{(subject, relation, object)} tuple.
    \item \(a\) is the correct answer derived from the object.
    \item \(c\) is an evidence passage intended to support or contradict \(a\).
\end{itemize}
\noindent
The model is given the pair \((q, c)\) and tasked with answering the question \(q\). Our goal is to evaluate whether it can correctly predict \(a\), even when the passage \(c\) contains misleading or conflicting information.

To probe the model’s robustness, we introduce targeted perturbations to the answer-bearing entity in \(c\) while keeping both \(q\) and \(a\) fixed. These perturbations are constructed at four increasing levels of contradiction—\emph{subtle}, \emph{mild}, \emph{moderate}, and \emph{blatant}—to test how the model balances contextual evidence against its parametric knowledge.

\paragraph{Illustrative Example.} In the \textsc{Wikipedia Years} domain, consider the question: \emph{“In which year was the census conducted that reported the population of Lukhi village in Iran as 35?”} The correct answer \(a\) is \texttt{2006}, based on the underlying knowledge tuple. We then modify the corresponding passage \(c\) to simulate contradiction:
\begin{itemize}[leftmargin=1.5em]
    \item A \textbf{moderate} perturbation replaces the year with \texttt{1966}, creating a subtle conflict.
    \item A \textbf{blatant} perturbation changes it to \texttt{2106}, introducing an implausible inconsistency.
\end{itemize}
The remainder of the passage remains unchanged. These graded interventions enable controlled analysis of how LLMs resolve factual conflicts under varying degrees of contextual reliability. More examples are put in Appendix~\ref{sec:examples}.

\subsection{Assessing Knowledge Recall}
\label{sec:knowledge}

To estimate whether a model \emph{knows} the correct answer \(a\) to a question \(q\) independently of any supporting passage, we employ a sampling-based probing method inspired by \textsc{SliCK}~\citep{gekhman2024does}. The core idea is to gauge how frequently a model recalls the correct answer when presented with the question alone, using no external context.

\paragraph{Sampling Procedure.}  
For each QA pair \((q, a)\), we generate $N=32$ completions from the base model at a fixed temperature \(T = 0.5\), using a 4-shot prompt for few-shot conditioning. Then a group of predictions is obtained as $\{\hat{a}_1, ..., \hat{a}_N\}$, and the \textit{sampled accuracy} is defined as:
\[
\mathrm{Acc}\left(q, a, \{\hat{a}_i\}_{i=1}^N\right) = \frac{1}{N} \sum_{i=1}^N \mathbbm{1} \, [\hat{a}_i = a],
\]
where \(\mathbbm{1}[\cdot]\) is the indicator function. This provides a robust estimate of the model’s prior knowledge of \(a\). This is distinct from \emph{self-confidence} (discussed later), which reflects belief in a single prediction.

\paragraph{Knowledge Categorization.}  
Based on sampled accuracy, we categorize each \((q, a)\) pair into one of four \textit{knowledge tiers}  (see Tab.~\ref{tab:knowledge_categories}):
\begin{itemize}[leftmargin=*]
    \item \textbf{Highly Known:} \(\mathrm{Acc}\left(q, a, \{\hat{a}_i\}_{i=1}^N\right)  \ge 0.85\) — strong and consistent recall,
    \item \textbf{Maybe Known:} \(0.30 \le \mathrm{Acc}\left(q, a, \{\hat{a}_i\}_{i=1}^N\right)  < 0.85\) — partial or inconsistent recall,
    \item \textbf{Weakly Known:} \(0 < \mathrm{Acc}\left(q, a, \{\hat{a}_i\}_{i=1}^N\right)  < 0.30\) — sporadic or weak recall,
    \item \textbf{Unknown:} \(\mathrm{Acc}\left(q, a, \{\hat{a}_i\}_{i=1}^N\right)  = 0\) — no evidence of prior recall.
\end{itemize}
This taxonomy allows us to stratify the model’s factual knowledge without relying on any auxiliary retrieval or context conditioning.

\subsection{Quantifying Self-Confidence}

\paragraph{Definition.} In addition to knowledge qualification, which reflects whether a model can recall a fact, we also measure how strongly the model \emph{believes} in its answer when generating a response. We define a model's \textbf{self-confidence} for a prediction \(\hat{a} = \langle t_1, \dots, t_L \rangle\) to question \(q\) as the average log-probability assigned to the answer tokens~\citep{wang2022self}:
\begin{equation*}
\label{equ:confidence}
    p_\theta(\hat{a}\mid q) = \frac{1}{L} \sum_{i=1}^{L} \log p(t_i \mid q, t_{<i}).
\end{equation*}
This score is derived from a single greedy prediction conditioned on a fixed few-shot prompt without temperature sampling. A higher confidence value implies greater internal belief in the generated answer.

\paragraph{Distinction from Knowledge Recall.}  
Unlike sampled accuracy, which captures the empirical frequency of correct recall across many completions, self-confidence reflects the strength of belief in a single response. These two axes—\textit{knowability} and \textit{belief strength}—may diverge: a model might confidently predict an incorrect answer (high confidence, low recall), or may inconsistently predict a correct one (low confidence, high recall).

\section{Method}
We introduce \textit{Self-Reflective Debate for Contextual Reliability (SR-DCR)}, a framework that combines asymmetric MAD with self-confidence–aware reasoning to evaluate context trustworthiness and determine final answers under conflicting information. SR-DCR unfolds in three stages:
\begin{enumerate}[leftmargin=*, itemsep=2pt]
    \item \textbf{Asymmetric Context Verification Debate (ACVD)}: Agents debate the reliability of the context from asymmetric viewpoints.
    \item \textbf{Self-Confidence Estimation}: The model predicts a zero-context answer \(\hat{a}_{\textsc{prior}}\) and computes its confidence score.
    \item \textbf{Final Answer Selection}: A decision rule chooses between \(\hat{a}_{\textsc{ctx}}\), \(\hat{a}_{\textsc{prior}}\), or \textsc{Abstain}, based on the context verdict and confidence level.
\end{enumerate}

\subsection{Asymmetric Context Verification Debate}
\label{subsec:acvd}

To assess the reliability of external context passages, we introduce \textit{ACVD}—a structured MAD framework that builds on prior MAD approaches~\citep{du2023improving,michael2023debate,lang2025debate}, but introduces informational asymmetry between agents. Unlike standard MAD setups where both agents receive the same input, ACVD withholds the context from one participant, enabling an adversarial test of whether the passage contributes trustworthy information.

We instantiate three roles:
\begin{itemize}[itemsep=1pt, topsep=2pt, leftmargin=*, labelsep=0.5em]
    \item \textbf{Defender (Agent A)} sees \((q, c)\) and defends the context-based answer \(\hat{a}_{\textsc{ctx}} := f_{\theta}(q, c)\), arguing that passage \(c\) is coherent and helpful.
    \item \textbf{Critic (Agent B)} sees only the question \(q\) and supports the prior answer \(\hat{a}_{\textsc{prior}}\), arguing that context \(c\) is misleading or fabricated.
    \item \textbf{Judge (Agent C)} observes the full debate transcript up to round \(r\), issuing a verdict \(\mathcal{V}^{(r)} \in \{\textsc{Reasonable}, \textsc{Unreasonable}\}\) based on the evolving dialogue.
\end{itemize}

The debate proceeds over \(R = 6\) rounds. In round \(r = 0\), both Defender and Critic submit opening statements. In subsequent rounds \(r \ge 1\), the Critic speaks first and the Defender replies. Each agent can access the full transcript \(\mathcal{T}^{(r)}\) up to that round and may quote or challenge prior arguments. The final verdict is defined as the earliest stabilized outcome:
\[
\mathcal{V}^* := \mathcal{V}^{(r^*)}, \, \text{where } \mathcal{V}^{(r)} = \mathcal{V}^{(r+1)} = \cdots = \mathcal{V}^{(R)}.
\]

ACVD allows us to assess not only whether models use context but whether they can recognize when context is verifiably beneficial or harmful—using a setup that foregrounds the asymmetry of information and the epistemic role of debate.

\subsection{Self-Confidence–Aware Belief Update}
 
Given a question \(q\), the model generates a prior prediction \(\hat{a}_{\textsc{prior}}\) under zero-context conditions. Let \(p_{\theta}(a \mid q)\) denote the model’s predictive distribution. The self-confidence score is defined as:
\[
\mathrm{Conf}(q, \hat{a}_{\textsc{prior}}) := p_{\theta}(\hat{a}_{\textsc{prior}} \mid q),
\]
computed via normalized log-probability over token sequences. We threshold this score at \(\tau = 0.90\):
\[
\textsc{Conf}(q) = 
\begin{cases}
\textsc{High}, & \text{if } \mathrm{Conf}(q, \hat{a}_{\textsc{prior}}) \ge \tau, \\
\textsc{Low}, & \text{otherwise}.
\end{cases}
\]
This binary confidence label is cached per instance and used to inform final decisions without repeated inference.

Finally, the model selects its answer using a gating decision that combines the verdict \(\mathcal{V}^{*}\) and confidence level:
\small
\[
\hat{a}_{\textsc{final}} = 
\begin{cases}
\hat{a}_{\textsc{ctx}}, & \text{if } \mathcal{V}^{*} = \textsc{Reasonable}, \\
\hat{a}_{\textsc{prior}}, & \text{if } \mathcal{V}^{*} = \textsc{Unreasonable}  \\ 
 & \land \textsc{Confidence}(q) = \textsc{High}, \\
\textsc{Abstain}, & \text{otherwise}.
\end{cases}
\]
\normalsize
This decision rule ensures that the model trusts external context only when it is explicitly judged to be reliable, and otherwise relies on its internal belief only if that belief is expressed with high confidence.

\begin{figure}[t]
  \centering
  \includegraphics[width=0.9\columnwidth]{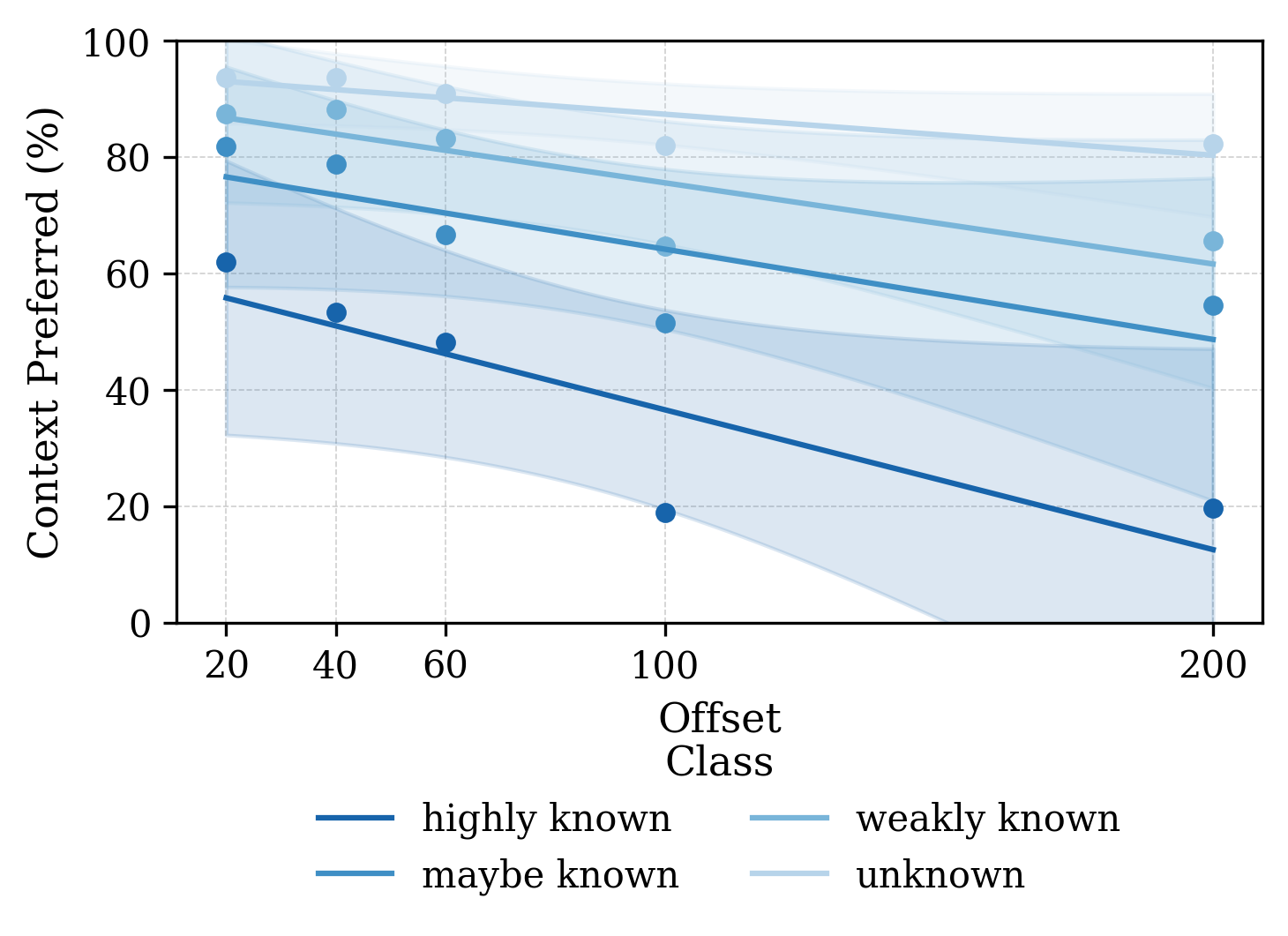}
  \vspace{-1em}
  \caption{Context preference rates of the few-shot baseline across perturbation levels and knowledge categories, where perturbation levels are represented by offset magnitudes: subtle (offset = 20), mild (40), moderate (60), and blatant (100 \& 200).}
  \label{fig:baseline-offset-pref}
  \vspace{-1em}
\end{figure}

\section{Experiments}

\paragraph{Dataset and Implementations.} We utilize \textit{ClashEval}~\cite{wu2024clasheval}, a benchmark specifically constructed to study factual conflicts between an LLM's prior knowledge and contradictory external context.  We adopt the official development/test splits and construct a controlled subset by sub-sampling and stratifying perturbation levels.  All models are evaluated using the \textit{exact match (EM)} metric: a prediction $\hat{a}$ is correct only if it exactly matches the ground truth $a$. More experimental details are in Appendix~\ref{sec:exp_details}, and additional results are available in Appendix~\ref{sec:addition_results}. 

\paragraph{Baselines.} Five strong baselines and the Golden Baseline are selected: 
(1) \emph{Few-shot prompting} provides the model with 4 randomly selected exemplars along with the context~\cite{brown2020language}. 
(2) \emph{Self-Ask}~\cite{press2022selfask} prompts the model to generate intermediate sub-questions before answering the main query. 
(3) \emph{RCI}~\cite{kim2023language} requires the model to iteratively critique and improve its output. 
(4) \emph{Judge Debate}~\cite{liang2023encouraging} classical MAD framework involves two debaters and a judge: one debater receives and supports the context-derived answer as initial stand, while the other supports prior-knowledge-derived answer as initial stand. 
(5) \emph{Naive Debate}~\cite{du2023improving} asks multiple LLMs to propose individual answers and engage in multi-round debate, ultimately converging on a final answer. (6) \emph{Golden Baseline} represents a theoretical upper bound in which models only trust the unperturbed, ground-truth context. When facing perturbed contexts, models rely entirely on their prior knowledge. This setup simulates an ideal scenario where LLMs perfectly discern context correctness and serves as a reference for their best achievable performance under our experimental setting.


\subsection{Few-Shot Context Reliance under Perturbation}
\label{subsec:baseline-offset}

We begin by investigating how a standard few-shot prompting strategy behaves in the presence of increasingly misleading context. We prompt the model with several random in-domain exemplars and present it with the input pair \((q, c)\), where \(c\) contains a perturbed version of the ground-truth answer. No adversarial interaction is used at this stage. For each example, we record whether the model's answer aligns with the (incorrect) contextual claim—i.e., whether it “prefers” the context over its prior knowledge. We then aggregate this context preference rate across different offset levels and stratify by knowledge category.

Fig.~\ref{fig:baseline-offset-pref} shows that \emph{Highly-Known} items rapidly reject perturbed context with increasing offsets. In contrast, \emph{Unknown} and \emph{Weakly Known} examples exhibit persistent reliance on context, remaining $>70\%$ agreement even under extreme perturbations. This overreliance on erroneous context in the absence of prior certainty motivates our SR-DCR design.

\subsection{Linking Self-Confidence and Prior Knowledge}

Next, we analyze the relationship between two measures of model familiarity: (i) sampling-based knowledge categories and (ii) single-shot self-confidence scores \(\mathrm{Conf}(q,\hat{a})\). We conduct this analysis across 5 LLMs: GPT-3.5-Turbo, GPT-4o~\citep{openai2023gpt4}, Claude Sonnet 3.7, Claude Haiku 3.5, and Llama 3.3-70B~\citep{meta2024llama3}.

For each model, we first compute \(C(\tau) = \{(q,a) \mid \mathrm{Conf}(q,\hat{a}) \ge \tau\}\) and define:
\small
\begin{align*}
P(\text{Highly-Known} \mid \mathrm{Conf} \ge \tau) = \frac{\left| S_{\mathrm{Highly}} \cap C(\tau) \right|}{\left| C(\tau) \right|}.
\end{align*}
\normalsize
where \(S_{\mathrm{Highly}}\) is the set of QA pairs categorized as Highly-Known.

\begin{figure}[t]
  \centering
  \includegraphics[width=\columnwidth]{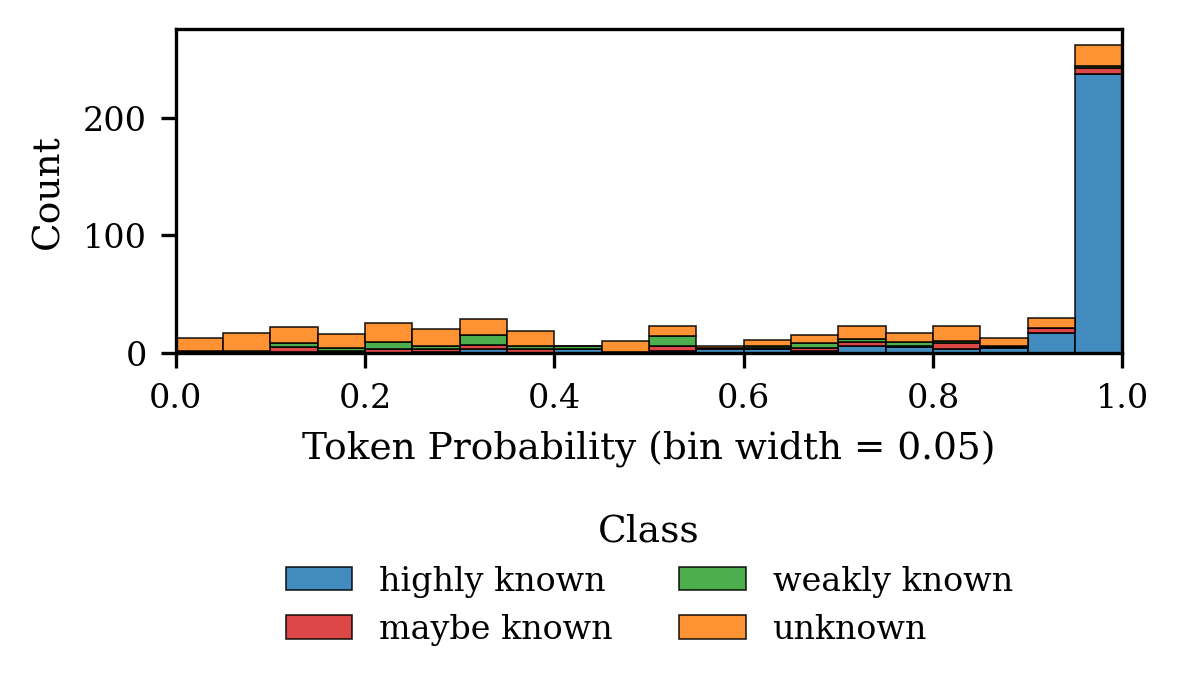}
  \vspace{-1em}
  \caption{
    The distribution of self-confidence scores of GPT-4o, colored by sampling-based knowledge categories.     Highly-Known instances cluster at the high end of the distribution (\(\ge 0.95\)), while Unknown examples dominate the low-confidence region, demonstrating self-confidence’s efficacy as a proxy for factual recall.
  }
  \label{fig:self confidence}
  \vspace{-1em}
\end{figure}

Fig.~\ref{fig:self confidence} and~\ref{fig:self confidence all} show that across all models, this conditional probability exceeds 0.88 once \(\tau \ge 0.90\) and surpasses 0.95 at \(\tau \ge 0.95\), confirming that high-confidence predictions are strong indicators of prior knowledge. In contrast, confidence below 0.70 yields poor overlap with Highly-Known pairs.

These findings validate our key hypothesis: when \(\mathrm{Conf}(q,\hat{a})\) crosses a threshold (e.g., 0.90), the model is highly likely to “know” the answer. We use this insight to guide dynamic routing in SR-DCR.
\begin{figure}[t]
  \centering
  \includegraphics[width=0.9\columnwidth]{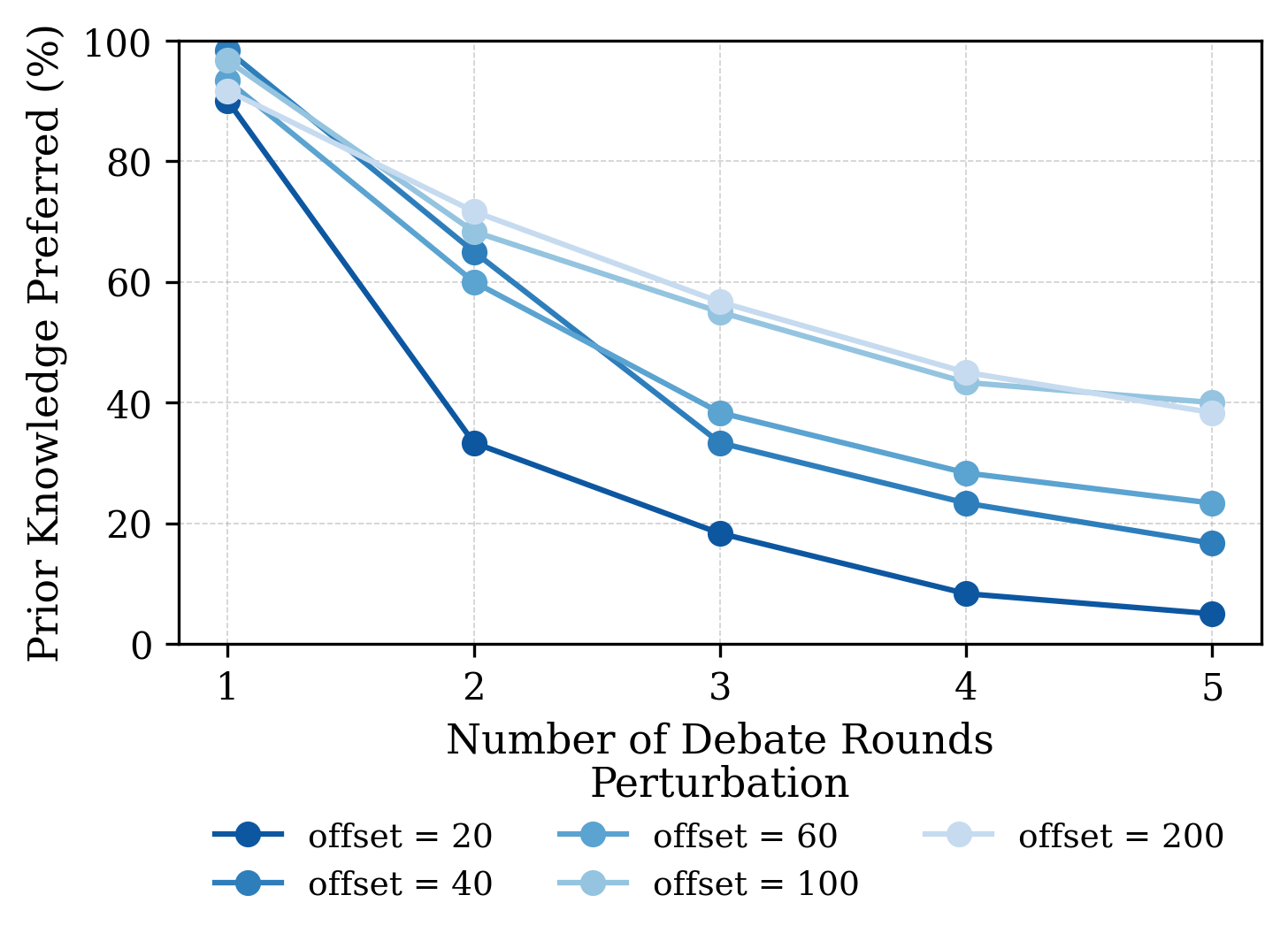}
  \caption{
    Prior Knowledge preference rate in judge debate on Sonnet 3.7 across debate rounds, stratified by perturbation levels (represented by offset). 
    Judges exhibit strong prior bias in early rounds, but shift toward context over rounds.
  }
  \vspace{-1em}
  \label{fig:acc vs offset}
\end{figure}

\subsection{Evaluating Multi-Agent Debate under Knowledge Conflict}
\label{subsec:mad-analysis}
\begin{table*}[!t]
  \centering
  \caption{Accuracy of each method across standard and perturbed contexts. Best results per column within each model block are \textbf{{bold}}, second-best are \underline{underlined}.}
  \label{tab:method-accuracy}
  \resizebox{0.65\textwidth}{!}{%
  \begin{tabular}{llccc} \toprule
    \textbf{Model} & \textbf{Method} & \textbf{Standard Context} & \textbf{Perturbed Context} & \textbf{Overall} \\ \midrule
    \multirow{9}{*}{GPT 3.5 Turbo} 
    & Few-Shots & \textbf{99.30\%} & 9.00\% & 54.15\% \\
    & Self-Ask & 95.00\% & 14.66\% & 54.83\% \\
    & RCI & 96.70\% & 12.32\% & 54.51\% \\
    & Judge Debate (Round 1) & 36.67\% & \underline{23.99\%} & 30.33\% \\
    & Judge Debate (Round 3) & 83.30\% & 14.02\% & 48.66\% \\
    & Judge Debate (Round 5) & 90.33\% & 17.00\% & 53.66\% \\
    & Naive Debate & \underline{98.33\%} & 20.33\% & \underline{59.33\%} \\
    & SR-DCR & 95.70\% & \textbf{29.66\%} & \textbf{62.68\%} \\
    \cmidrule(lr){2-5}
    & Golden Baseline & 98.67\% & 31.67\% & 65.17\% \\
    \midrule
    \multirow{9}{*}{GPT 4o}
    & Few-Shots & \textbf{99.33\%} & 42.27\% & 70.80\% \\
    & Self-Ask & 97.00\% & 31.67\% & 64.33\% \\
    & RCI & 91.67\% & 22.33\% & 57.00\% \\
    & Judge Debate (Round 1) & 70.00\% & \textbf{57.33\%} & 63.67\% \\
    & Judge Debate (Round 3) & 96.00\% & 45.67\% & \underline{70.83\%} \\
    & Judge Debate (Round 5) & 96.00\% & 45.33\% & 70.67\% \\
    & Naive Debate & \underline{97.33\%} & 44.33\% & \underline{70.83\%} \\
    & SR-DCR & 94.67\% & \underline{54.51\%} & \textbf{74.59\%} \\
    \cmidrule(lr){2-5}
    & Golden Baseline & 99.00\% & 50.00\% & 74.50\% \\
    \midrule
    \multirow{9}{*}{Claude Haiku 3.5}
    & Few-Shots & \textbf{99.67\%} & 16.33\% & 58.00\% \\
    & Self-Ask & 98.00\% & 21.67\% & 59.83\% \\
    & RCI & 93.33\% & 22.00\% & 57.66\% \\
    & Judge Debate (Round 1) & 55.67\% & 26.33\% & 41.00\% \\
    & Judge Debate (Round 3) & 94.67\% & 21.67\% & 58.17\% \\
    & Judge Debate (Round 5) & 97.00\% & 21.00\% & 59.00\% \\
    & Naive Debate & \underline{98.67\%} & \underline{29.67\%} & \underline{64.17\%} \\
    & SR-DCR & 95.33\% & \textbf{49.89\%} & \textbf{72.61\%} \\
    \cmidrule(lr){2-5}
    & Golden Baseline & 99.67\% & 48.00\% & 73.83\% \\
    \midrule
    \multirow{9}{*}{Claude Sonnet 3.7}
    & Few-Shots & \underline{98.67\%} & 43.04\% & 70.85\% \\
    & Self-Ask & 97.00\% & 45.33\% & \underline{71.17\%} \\
    & RCI & 95.00\% & 22.66\% & 58.83\% \\
    & Judge Debate (Round 1) & 61.67\% & \textbf{50.33\%} & 56.00\% \\
    & Judge Debate (Round 3) & \underline{98.67\%} & 30.00\% & 64.33\% \\
    & Judge Debate (Round 5) & \textbf{99.33\%} & 24.67\% & 62.00\% \\
    & Naive Debate & 98.00\% & 47.34\% & \textbf{72.67\%} \\
    & SR-DCR & 91.04\% & \underline{47.94\%} & 69.76\% \\
    \cmidrule(lr){2-5}
    & Golden Baseline & 99.67\% & 47.67\% & 73.67\% \\
    \midrule
    \multirow{9}{*}{Llama 3.3-70B}
    & Few-Shots & \textbf{98.67\%} & 9.67\% & 54.17\% \\
    & Self-Ask & \underline{98.33\%} & 23.67\% & \underline{61.00\%} \\
    & RCI & 94.96\% & 25.00\% & 59.98\% \\
    & Judge Debate (Round 1) & 56.67\% & 35.66\% & 46.17\% \\
    & Judge Debate (Round 3) & 92.33\% & \textbf{28.33\%} & 60.33\% \\
    & Judge Debate (Round 5) & 93.33\% & 25.67\% & 59.49\% \\
    & Naive Debate & \textbf{98.67\%} & 22.33\% & 60.50\% \\
    & SR-DCR & 94.97\% & \underline{28.39\%} & \textbf{61.67\%} \\
    \cmidrule(lr){2-5}
    & Golden Baseline & 99.00\% & 33.33\% & 66.17\% \\
    \bottomrule
  \end{tabular}
  }
  \vspace{-0.5em}
\end{table*}

\begin{table*}[ht]
  \centering
  \caption{Performance of different methods under a group of prevalent LLMs across varying offset levels. Best and second-best results are shown in \textbf{bold} and \underline{underlined}, respectively.}
  \label{tab:accuracy-offset}
  \resizebox{0.75\textwidth}{!}{%
  \begin{tabular}{llrrrrrr}
    \toprule
    \textbf{Model} & \textbf{Method} & \textbf{Offset 20} & \textbf{Offset 40} & \textbf{Offset 60} & \textbf{Offset 100} & \textbf{Offset 200} & \textbf{Overall} \\
    \midrule
    \multirow{8}{*}{GPT 3.5 Turbo}
     & Few-Shots & 1.70 & 8.30 & 11.70 & 8.30 & 15.00 & 9.00 \\
     & Self-Ask & 1.00 & 13.30 & 18.30 & 23.30 & 18.30 & 14.66 \\
     & RCI & 8.30 & 8.30 & 15.00 & 16.70 & 13.30 & 12.32 \\
     & Judge Debate (Round 1) & \underline{25.00} & \underline{21.67} & \underline{28.30} & 18.33 & 26.67 & \underline{23.99} \\
     & Judge Debate (Round 3) & 5.00 & 11.70 & 20.00 & 11.70 & 21.70 & 14.02 \\
     & Judge Debate (Round 5) & 10.00 & 13.33 & 16.67 & 20.00 & 25.00 & 17.00 \\
     & Naive Debate & 8.33 & 13.33 & 15.00 & \underline{25.00} & \textbf{40.00} & 20.33 \\
     & SR-DCR & \textbf{26.70} & \textbf{25.00} & \textbf{30.00} & \textbf{28.30} & \underline{38.30} & \textbf{29.66} \\
    \cmidrule(lr){2-8}
     & Golden Baseline & 31.67 & 30.00 & 33.33 & 21.67 & 41.67 & 31.67 \\
    \midrule
    \multirow{8}{*}{GPT 4o}
     & Few-Shots & 23.33 & 31.00 & 38.33 & \underline{57.00} & 61.67 & 42.27 \\
     & Self-Ask & 28.33 & 25.00 & 21.67 & 40.00 & 43.33 & 31.67 \\
     & RCI & 15.00 & 16.67 & 21.67 & 23.33 & 35.00 & 22.33 \\
     & Judge Debate (Round 1) & \textbf{56.67} & \textbf{53.33} & \textbf{58.33} & \textbf{58.33} & 60.00 & \textbf{57.33} \\
     & Judge Debate (Round 3) & 35.00 & 45.00 & 48.33 & 48.33 & 51.67 & 45.67 \\
     & Judge Debate (Round 5) & 33.33 & 41.67 & 48.33 & 48.33 & 55.00 & 45.33 \\
     & Naive Debate & 26.67 & 33.33 & 36.67 & 53.33 & \textbf{71.67} & 44.33 \\
     & SR-DCR & \underline{47.46} & \underline{47.37} & \underline{52.73} & 55.00 & \underline{70.00} & \underline{54.51} \\
    \cmidrule(lr){2-8}
     & Golden Baseline & 50.00 & 50.00 & 53.33 & 46.67 & 50.00 & 50.00 \\
    \midrule
    \multirow{8}{*}{Claude Haiku 3.5}
     & Few-Shots & 8.33 & 18.33 & 6.67 & 23.33 & 25.00 & 16.33 \\
     & Self-Ask & 10.00 & 16.67 & 25.00 & 25.00 & 31.67 & 21.67 \\
     & RCI & 8.33 & 13.33 & 15.00 & 30.00 & 43.33 & 22.00 \\
     & Judge Debate (Round 1) & \underline{28.33} & \underline{26.67} & 25.00 & 18.33 & 33.33 & 26.33 \\
     & Judge Debate (Round 3) & 15.00 & 20.00 & 23.33 & 20.00 & 30.00 & 21.67 \\
     & Judge Debate (Round 5) & 15.00 & 15.00 & 21.67 & 21.67 & 31.67 & 21.00 \\
     & Naive Debate & 18.33 & 21.67 & \underline{26.67} & \underline{35.00} & \underline{46.67} & \underline{29.67} \\
     & SR-DCR & \textbf{55.56} & \textbf{41.38} & \textbf{53.45} & \textbf{47.27} & \textbf{51.79} & \textbf{49.89} \\
    \cmidrule(lr){2-8}
     & Golden Baseline & 50.00 & 41.67 & 55.00 & 43.33 & 50.00 & 48.00 \\
     \midrule
     \multirow{8}{*}{Claude Sonnet 3.7}
     & Few-Shots & 26.67 & 31.67 & 38.33 & 55.00 & 63.52 & 43.04 \\
     & Self-Ask & 25.00 & 35.00 & 41.67 & \underline{63.33} & 61.67 & 45.33 \\
     & RCI & 13.33 & 18.33 & 20.00 & 23.33 & 38.33 & 22.66 \\
     & Judge Debate (Round 1) & \textbf{46.67} & \textbf{43.33} & \textbf{56.67} & 48.33 & 56.67 & \textbf{50.33} \\
     & Judge Debate (Round 3) & 15.00 & 18.33 & 30.00 & 41.67 & 45.00 & 30.00 \\
     & Judge Debate (Round 5) & 3.33 & 10.00 & 23.33 & 40.00 & 46.67 & 24.67 \\
     & Naive Debate & 21.67 & 31.67 & \underline{45.00} & \textbf{66.67} & \textbf{71.67} & 47.34 \\
     & SR-DCR & \underline{33.33} & \underline{40.00} & 43.63 & 55.00 & \underline{68.42} & \underline{47.95} \\
    \cmidrule(lr){2-8}
     & Golden Baseline & 50.00 & 43.33 & 53.33 & 41.67 & 50.00 & 47.67 \\
    \midrule
     \multirow{8}{*}{Llama 3.3-70B}
     & Few-Shots & 5.00 & 3.33 & 13.33 & 11.67 & 15.00 & 9.67 \\
     & Self-Ask & 15.00 & 18.33 & 21.67 & 28.33 & 35.00 & 23.67 \\
     & RCI & 13.33 & 18.33 & 25.00 & 31.67 & 36.67 & 25.00 \\
     & Judge Debate (Round 1) & \textbf{35.00} & \textbf{40.00} & \textbf{33.33} & 28.33 & \textbf{41.67} & \textbf{35.67} \\
     & Judge Debate (Round 3) & 16.67 & 18.33 & \underline{31.67} & \textbf{35.00} & \underline{40.00} & 28.33 \\
     & Judge Debate (Round 5) & 15.00 & 15.00 & \textbf{33.33} & 31.67 & 33.33 & 25.67 \\
     & Naive Debate & 11.67 & 13.33 & 20.00 & 31.67 & 35.00 & 22.33 \\
     & SR-DCR & \underline{20.34} & \underline{25.00} & 30.51 & \underline{32.76} & 33.33 & \underline{28.39} \\
    \cmidrule(lr){2-8}
     & Golden Baseline & 50.00 & 41.67 & 55.00 & 43.33 & 50.00 & 48.00 \\
    \bottomrule
  \end{tabular}
  }
\vspace{-1em}
\end{table*}

We compare the Classical MAD framework (Judge Debate) against five baselines with standard and perturbed contexts. Results of 1, 3, 5 rounds of debate on five different models are shown in Tab.~\ref{tab:method-accuracy}, from which we draw several conclusions. 

\paragraph{Insight 1 — Debate rounds shift judgment from prior to context.}  
We observe a consistent trend in which early rounds of judge debate favor prior knowledge answers, while longer debates increasingly support context-derived responses. At \(r = 1\), judges tend to reject the context. With GPT-4o, judge debate (\(r = 1\)) achieves only 70\% accuracy on standard (unperturbed) context questions, compared to $>90\%$ with other baselines. This skepticism toward context provides strong protection against misinformation: on perturbed inputs, GPT-4o’s judge debate  (\(r = 1\)) achieves significantly higher accuracy (57.33\%) than others, with Self-Ask and RCI trailing far behind at 31.7\% and 22.3\%, respectively.

As the number of debate rounds increases, this behavior gradually reverses. Judges become more willing to trust contextual evidence—especially when it is accurate. On standard (unperturbed) examples, accuracy rises steadily with more rounds, reaching nearly 100\% by \(r = 3\). This benefit is not without tradeoffs. While longer debates improve acceptance of correct context, they also reduce robustness to misleading information. accuracy on perturbed inputs declines as the number of rounds increases: for GPT-4o, accuracy on perturbations drops from 57.3\% at \(n = 1\) to 45.3\% at \(n = 5\).

\paragraph{Insight 2 — Longer debates increase context reliance, and reduce robustness to small inconsistencies.}  

A more detailed analysis by perturbation level reveals that this degradation is not uniform. From Fig.~\ref{fig:acc vs offset} and Fig.~\ref{fig:acc vs offset all}, we can see that for GPT-4o, sonnet 3.7, and LLaMA 70B, data points with a lower perturbation level (e.g., subtle (offset \(= 20\))) show a steeper decrease in accuracy, suggesting that minor inconsistencies are more easily overlooked as debates grow longer. In contrast, large perturbations (e.g., blatant (offset \(>= 100\))) remain consistently detectable: Their accuracy remains relatively stable across all debate rounds. This indicates that judge debate is effective in rejecting an obviously flawed context but becomes more vulnerable to subtle misinformation as it attempts to reconcile competing claims.

These observations reveal a tension in Judge Debate: short debates reinforce prior bias, while longer debates risk accepting false context. 


\subsection{Evaluation of SR-DCR}
\label{sec:sc_evaluation}

\paragraph{Evaluation Setups.} We assembled a composite testbed of 600 question–answer instances:  a half use \emph{standard contexts}, where the retrieved passage is correct; the other half employ \emph{perturbed contexts}, simulating erroneous RAG outputs at 4 calibrated disturbance levels (subtle to blatant perturbations).  This split measures the ability to leverage valid context and the robustness to misleading excerpts.



\paragraph{Results.}  Tab.~\ref{tab:method-accuracy} reports EM across standard and perturbed contexts. Across all model families, we observe a consistent pattern: few-shot prompting and retrieval-based approaches (e.g., Self-Ask, RCI) perform well on standard contexts but degrade sharply under adversarial perturbation. For instance, GPT-3.5 Turbo's accuracy drops from 99.3\% to 9.0\% with Few-Shot, and from 95.0\% to 14.7\% with Self-Ask. Classical MAD (e.g., Judge Debate at Round 5) improves robustness but often sacrifices standard-context accuracy. For example, Judge Debate (R5) recovers perturbed accuracy to 17.0\%, but standard accuracy falls to 90.3\%.

{SR-DCR} consistently outperforms prior methods in handling conflicting information. On GPT-3.5 Turbo, it achieves 29.7\% on perturbed inputs—nearly matching the golden baseline’s 31.7\%—while maintaining 95.7\% on standard contexts. This yields an overall accuracy of 62.7\%, a +3.4-point gain over Naive Debate, the best-performing baseline. On stronger models like GPT-4o and Haiku 3.5, SR-DCR matches or exceeds the performance of Naive Debate and classical MAD, with the largest gains observed in perturbed contexts: +10.2 points over Judge Debate (R5) on GPT-4o and +20.2 points over RCI on Haiku 3.5.

To better understand robustness under varying degrees of contradiction, Tab.~\ref{tab:accuracy-offset} reports accuracy across perturbation offsets. On GPT-3.5 Turbo, SR-DCR achieves state-of-the-art performance at every offset level, including 26.7\% at Offset 20 and 38.3\% at Offset 200—far surpassing all baselines. Similarly, SR-DCR attains the best average perturbed accuracy on Sonnet (47.9\%) and Haiku (49.9\%), and ranks second only to Judge Debate (R1) on GPT-4o. This demonstrates SR-DCR’s capacity to retain valid contextual information while resisting misleading perturbations.

We further analyze SR-DCR's ability to selectively trust valid contexts and reject corrupted ones. On GPT-3.5 Turbo, SR-DCR yields a true positive rate (standard context accuracy) of 95.7\% and a true negative rate (perturbed context accuracy) of 29.7\%. In contrast, Naive Debate achieves a higher standard-context rate (98.3\%) but a lower rejection rate (20.3\%), while Judge Debate (R5) performs worse on both fronts. SR-DCR thus achieves a more balanced trade-off, leveraging external context when warranted and falling back on internal knowledge when necessary.

\paragraph{Summary and Analysis.}  

Empirically, this self-reflective control yields three key advantages: (i) a +7.7-point improvement in robustness to corrupted context, (ii) a +5.6-point recovery in accuracy on clean, context-consistent cases, and (iii) a 2× reduction in prior bias during extended debates. Crucially, SR-DCR achieves this improved arbitration between parametric and contextual knowledge with minimal overhead, requiring only one additional forward pass for confidence estimation and a debate that is already necessary for context evaluation.

Together, these results confirm SR-DCR as a principled and efficient solution for resolving factual conflicts, outperforming both confidence-based and debate-only strategies across models and perturbation settings.



    
     %
\section{Related Work}


\paragraph{Knowledge Conflict.} In in-context learning, conflicts between an LLM’s internal knowledge and external context can cause interference~\citep{jin-etal-2024-tug}. Models often over-rely on coherent external evidence even when it contradicts their memory~\citep{xie2024adaptive}, especially under low internal confidence~\citep{xu-etal-2024-earth,chatziveroglou2025exploringllmreasoningcontrolled}. To address this, prior work has proposed Knowledge-Aware Fine-Tuning~\citep{li-etal-2023-large}, opinion-based prompts and counterfactuals~\citep{zhou-etal-2023-context}, and fact duration prediction~\citep{zhang-choi-2023-mitigating}. Some prioritize context~\citep{li-etal-2023-large,zhou-etal-2023-context,zhang-choi-2023-mitigating}, while others favor memory~\citep{hong-etal-2024-gullible}. However, rigid faithfulness to either source is suboptimal. We extend this line by jointly varying perturbation strength and prior confidence to better arbitrate between conflicting knowledge sources.

\paragraph{Multi-Agent Debate.} MAD has been shown to improve LLM reasoning and evaluation~\citep{khan2024debating}. Adversarial critique can enhance factuality, and aggregation strategies like RECONCILE~\citep{chen-etal-2024-reconcile} use confidence-weighted voting across agents. Prior work demonstrates that adversarial dialogue can reveal false assumptions~\citep{du2023improving}, improve robustness~\citep{michael2023debate}, and enhance factual accuracy even with weaker judges~\citep{kenton2024scalable}. However, most MAD frameworks debate all queries uniformly, ignoring model confidence—sometimes leading to unnecessary correction when the model is already correct. To address this, we integrate self-confidence signals with context reasonableness and introduce a judge agent to yield more balanced and credible answers.

\section{Conclusion}
We presented SR-DCR, a self-reflective framework that combines model confidence with asymmetric debate to resolve conflicts between parametric knowledge and contextual input. By selectively trusting context when it is judged reliable—and deferring to priors when confidence is high—SR-DCR improves factual robustness and accuracy under perturbation, outperforming classical debate and confidence-only baselines. Limitations are in Appendix~\ref{sec:limitations}.

\bibliography{custom}

\appendix
\newpage 

\section{Experimental Details}
\label{sec:exp_details}
\subsection{Base Models and Dataset}
Our experiments were conducted using a variety of state-of-the-art language models sourced from different platforms. From the Anthropic API, we used two versions of Claude models: \texttt{claude-3-5-haiku-20241022} and \texttt{claude-3-7-sonnet-20250219}. From OpenAI's API platform, we employed \texttt{gpt-3.5-turbo-0125} and the more recent \texttt{gpt-4o-2024-08-06}. For the LLaMA models, we used the \texttt{LLaMA 3.1 8B}, \texttt{LLaMA 3.3 70B}, and \texttt{LLaMA 3.1 405B}. 

To support experiments with local LLaMA models, we used three NVIDIA H100 GPUs. Running a full 5-round, 3-agent debate setup over our 600-question dataset required approximately 7 hours of wall-clock time on this setup. This runtime includes both the multi-agent dialogue stages and final judge evaluations.

The benchmark dataset, ClashEval, spans various knowledge domains—including biographical facts, geography, medical dosages, Olympic records, historical data, and breaking news—ensuring wide coverage in both content and difficulty.

\subsection{Baseline Implementation Details}
We implemented six baseline methods alongside SR-DCR to evaluate the effectiveness of different prompting and reasoning strategies for answering factual questions under a potentially perturbed context. Each method was run under standardized conditions across models and datasets to ensure fair comparison. Unless otherwise stated, all experiments used full-precision models with temperature set to zero and default decoding parameters.

\paragraph{Few-shot Prompting:} We provided the model with four randomly selected in-domain exemplars, along with the question and accompanying context and short instruction. 

\paragraph{Self-Ask~\cite{press2022selfask}:} It is designed to encourage structured intermediate reasoning by prompting the model to first generate a clarifying sub-question, answer that sub-question, and then commit to a final answer. We adopt a fixed template to ensure consistency across examples. The model is instructed to respond in exactly three lines—Sub-question, Sub-answer, and Final answer—with no additional output and make a best-guess answer even when uncertain. To ensure that the model fully completes all three required lines of output, we set the maximum token limit to 400 during decoding. 

\paragraph{RCI (Recursive Criticism and Improvement)~\cite{kim2023language}:} It encourages iterative self-reflection and refinement by prompting the model to first generate an initial answer, then critique that answer, and finally revise it based on its critique. This process simulates a single-agent self-debate loop. Our implementation follows a three-step pipeline. First, the model is prompted with the question and context and asked to produce its best guess (initial answer). Next, it receives the same input along with its previous response and is asked to identify any factual or logical issues in the form of a short, bulleted critique. Finally, the model is prompted to revise its answer using the critique, outputting only the updated final answer. Each of these three stages can response with a maximum of 300 tokens to ensure sufficient space for generation.

\paragraph{Naive Debate:} We use three independent agents of the same language model, all operating with a temperature of 0 and a maximum token limit of 250.  The agents engage in a maximum of five debate rounds to iteratively refine their answers. Each agent begins by independently reading the context and question and giving its best answer.

In each subsequent round, agents are shown their peers' answers from previous rounds. They are prompted to (1) evaluate the plausibility of those answers, (2) reflect on their reasoning, and (3) decide whether to revise their answer or stick with their original. The agents are explicitly instructed to avoid vague responses or uncertainty and to always produce a single answer. The debate terminates early if all three agents converge on the same answer. If no consensus is reached after five rounds, we adopt a majority vote to determine the final output.

\paragraph{Classical Debate \& Judge Debate:} We simulate a debate between three agents instantiated from the same language model, using a temperature of 0 and a maximum token limit of 300. Before the debate starts, we first obtain two initial answers: one is based on context, the other is purely based on prior knowledge. Two agents serve as debaters (Agent A and Agent B), and the third acts as a neutral judge. The debaters are initialized with the same instructions but different initial standings. Specifically, in our \emph{Judge Debate} setting, Agent A is instructed to support and defend an answer grounded in the context. In contrast, Agent B is instructed to support the answer based on prior knowledge.

Following their initial responses, the two agents engage in a multi-round debate where they alternate presenting arguments, challenging each other's positions, and defending their answers. After 1, 3, or 5 rounds of debate, the judge agent is presented with the question, context, and complete debate transcript and tasked with selecting a winning answer. The judge's final output must include a single answer followed by a concise justification. 

\begin{table*}[th]
  \centering
  \caption{Knowledge qualification categories based on the sampled recall rate from 32 completions at \(T = 0.5\). This categorization adapts the schema from \textsc{SliCK}~\citep{gekhman2024does}.}
  \label{tab:knowledge_categories}
  \resizebox{\textwidth}{!}{%
    \begin{tabular}{@{} l | l |c |p{7cm} @{}}
      \toprule
      \textbf{Type} 
        & \textbf{Category} 
        & \textbf{Definition} 
        & \textbf{Intuition} \\
      \midrule
      \multirow{3}{*}{Known}
        & Highly Known 
        & $\mathrm{P_{\!correct}}(q,a;M,T>0) \ge 0.85$ 
        & Model almost always recalls the fact. \\
        & Maybe Known 
        & $0.30 \le \mathrm{P_{\!correct}}(q,a;M,T>0) < 0.85$ 
        & Fact is recalled, but inconsistently. \\
        & Weakly Known       
        & $0 < \mathrm{P_{\!correct}}(q,a;M,T>0) < 0.30$ 
        & The model shows only sporadic recall. \\ \midrule 
      \addlinespace
      Unknown 
        & Unknown 
        & $\mathrm{P_{\!correct}}(q,a;M,T>0) = 0$ 
        & Model never recalls the fact. \\
      \bottomrule
    \end{tabular}%
  }
\end{table*}
\subsection{Knowledge Categories} 
We depict the details of knowledge categories in Tab.~\ref{tab:knowledge_categories}, where four types are included based on the sampled accuracy.

\section{More Illustrative Examples}
\label{sec:examples}
Consider the ClashEval question “Which city hosted the 1904 Summer Olympics?” with ground‑truth \emph{St.\,Louis}.  When prompted with four random exemplars and sampled 32 times at $T=0.5$, the model answers “St.\,Louis” in 28 trials, yielding a sampled accuracy of $0.875$ and a \emph{Highly‑Known} label~\cite{wu2024clasheval}.  In contrast, the question “Which team won the 1974 World Series?” might yield only 6 correct out of 32 ($\approx0.19$), classifying it as \emph{Weakly‑Known}.

\begin{figure*}
    \centering
    \includegraphics[width=\textwidth]{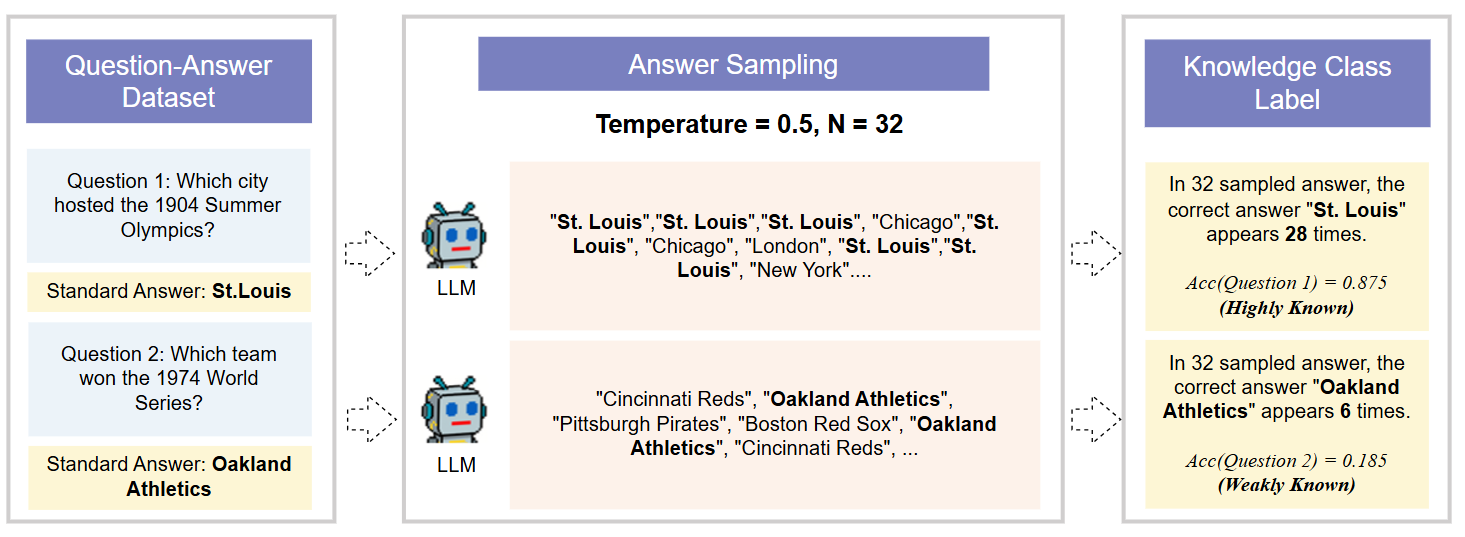}
    \caption{Two illustrative examples of how questions are classified based on the model's familiarity}
    \label{fig:example}
\end{figure*}

As shown Fig.~\ref{fig:example}, this illustrates how sampling with random four‑shot prompts uncovers differing degrees of model familiarity.

\section{Additionial Results}
\label{sec:addition_results}

\begin{figure*}[t]
    \centering
    \includegraphics[width=\textwidth]{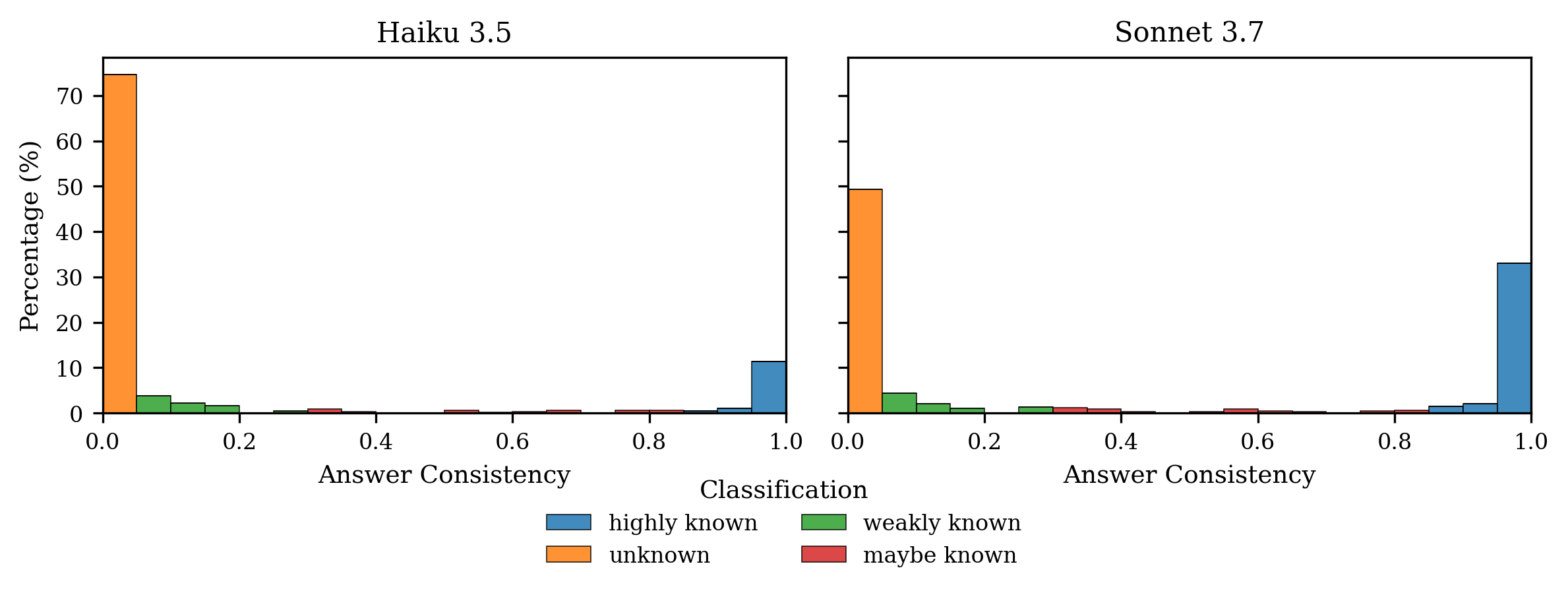}
    \caption{Distribution of self-consistency scores (estimated self-confidence score based on sampling) on our dataset, colored by sampling-based knowledge categories.}
    \label{fig:self consistency}
\end{figure*}

\subsection{Verification of Self-consistency Scores} 

For closed-source models that do not expose token-level log-probabilities (e.g., Anthropic's Claude models), we approximate model self-confidence using a sampling-based self-consistency measure. Specifically, we first query the model using few-shot prompt without context to obtain a deterministic answer ($T = 0$). Then, we sample the same question \textbf{16 times} using a higher temperature setting ($T = 0.5$). We compute the proportion of sampled outputs that match the deterministic answer, which we interpret as the model’s \emph{self-consistency} on that question. This score serves as a proxy for the model’s confidence in its original answer: higher consistency indicates higher certainty. 

The distribution of self-consistency scores using a stacked histogram is shown in Fig.~\ref{fig:self consistency}, which shows a similar pattern to the self-confidence scores derived from models with access to log-probabilities. This alignment supports the effectiveness of self-consistency as an approximation for model confidence, proving that our framework is also effective on closed-source models.

\subsection{More Experiment Results}

Fig.~\ref{fig:self consistency} presents the distribution of self-consistency scores (our proxy for self-confidence) for GPT-3.5 Turbo, GPT-4o, LLaMA 3.1 405B, and LLaMA 3.3 70B, extending the analysis shown in Fig.~\ref{fig:self consistency}. All four models exhibit similar self-confidence patterns, supporting the generality and robustness of our sampling-based approximation method.

\begin{figure*}
    \centering
    \includegraphics[width=\textwidth]{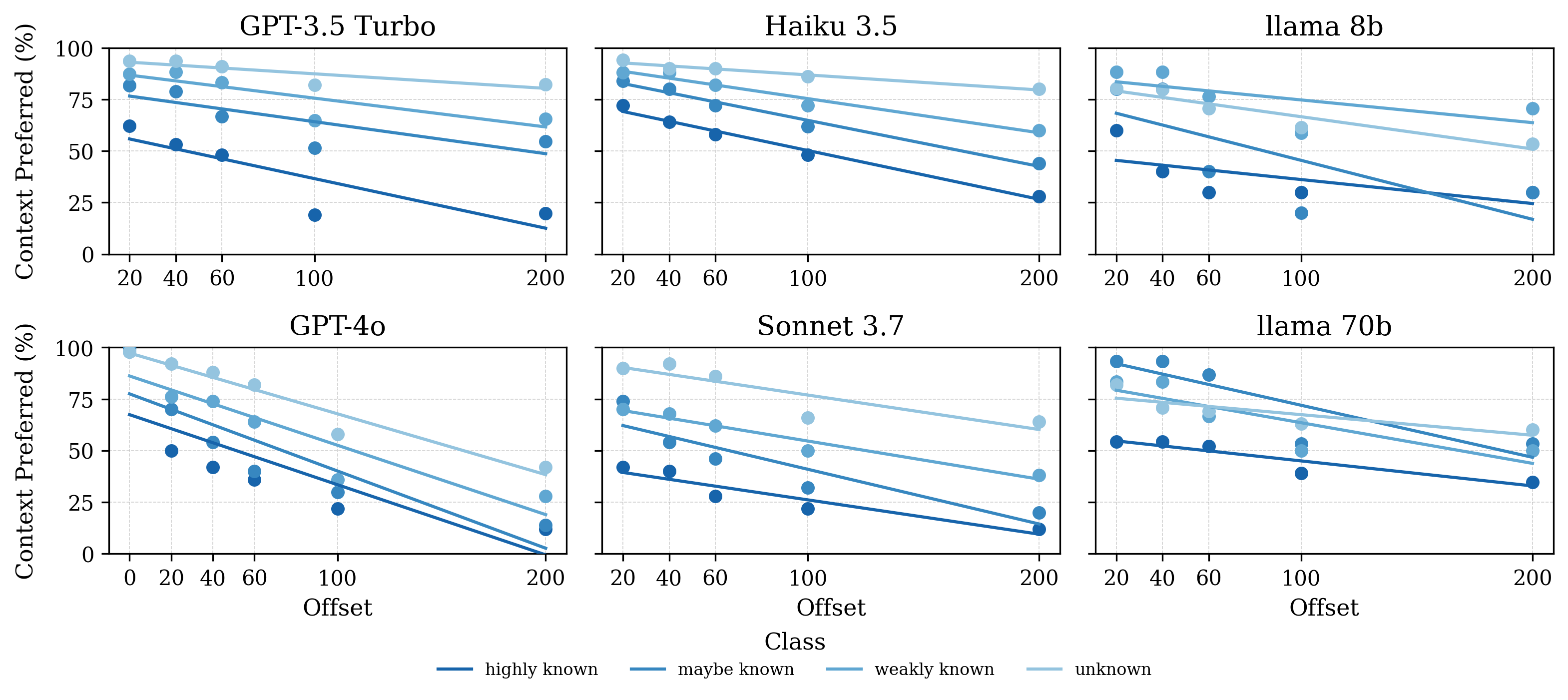}
    \caption{Context preference rates of a 4-shot baseline across offset magnitudes and knowledge categories of six different models on our dataset. Each curve indicates the proportion of examples where the model’s answer aligns with the (incorrect) perturbed context.}
    \label{fig:Context preference all}
\end{figure*}

Fig.~\ref{fig:Context preference all} extends our earlier analysis of the relationship between context preference and knowledge familiarity to six language models, including both open and closed-source LLMs. Each curve represents how often a model aligns its answer with the perturbed (and incorrect) context across varying offset magnitudes, stratified by knowledge category. The consistent trends across models—higher context preference in "unknown" or "unsure" cases, and lower preference in "highly known" cases—further validate the generality of our framework.

\begin{figure*}
    \centering
    \includegraphics[width=\textwidth]{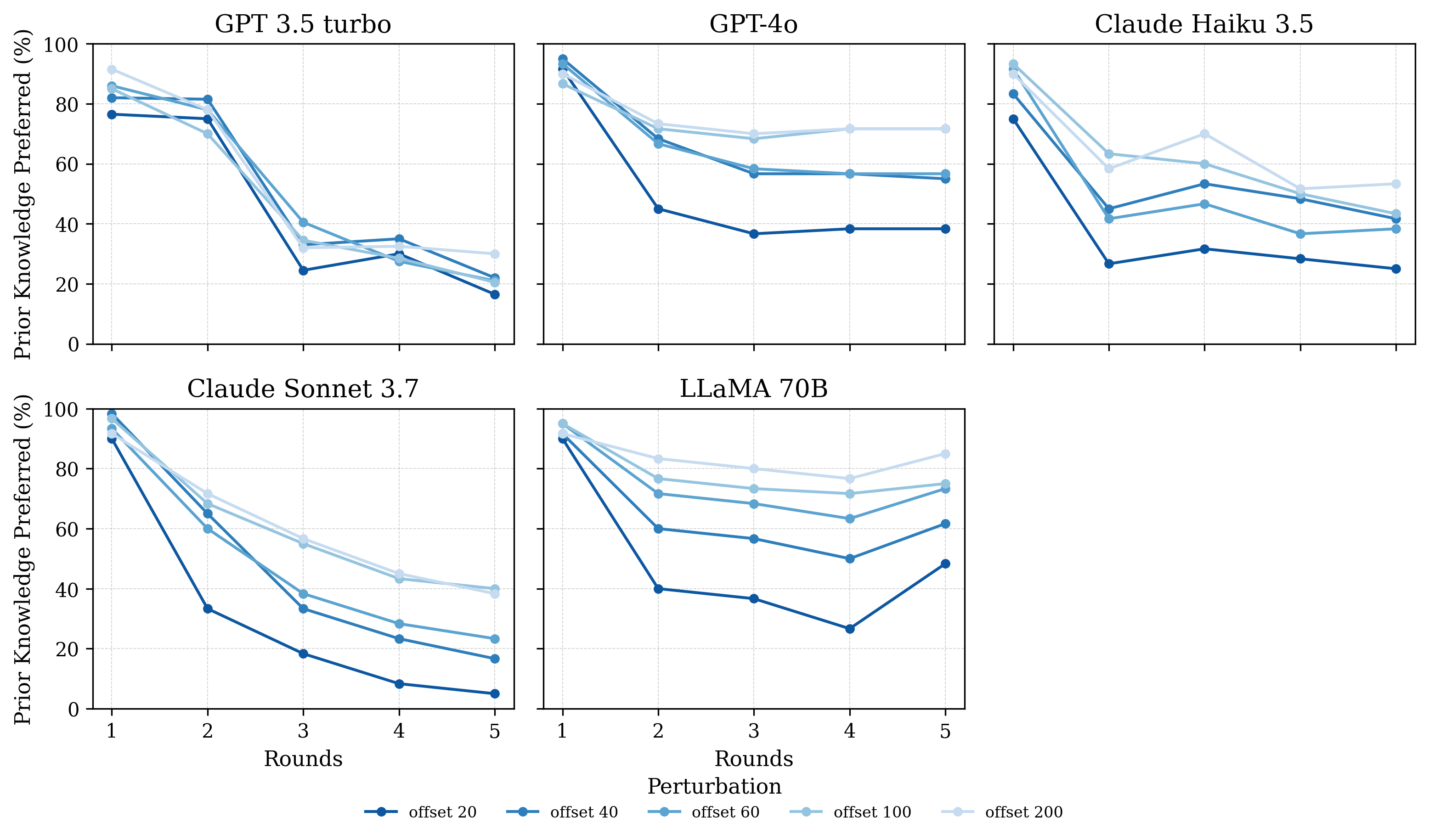}
    \caption{Prior Knowledge preference rate in judge debate of five models on perturbed context across debate rounds, stratified by five levels of perturbation strengths—subtle (offset = 20), mild (40), moderate (60), and blatant (100 \& 200).}
    \label{fig:acc vs offset all}
\end{figure*}

Fig.~\ref{fig:acc vs offset all} extends the analysis of context susceptibility by visualizing prior knowledge preference rates across five models, stratified by perturbation strength. Complementing the trends shown in Fig.~\ref{fig:acc vs offset}, this figure further demonstrates that models are more likely to accept the perturbed context as the debate rounds increase. Subtle inconsistencies (e.g., offset = 20) often lead to higher context preference after 3 rounds. For GPT 3.5 Turbo and Sonnet 3.7, the Prior knowledge preference rate is even below 20\% at the end of the debate when provided with context with minor error. 

\begin{figure*}[t]
    \centering
    \includegraphics[width=\textwidth]{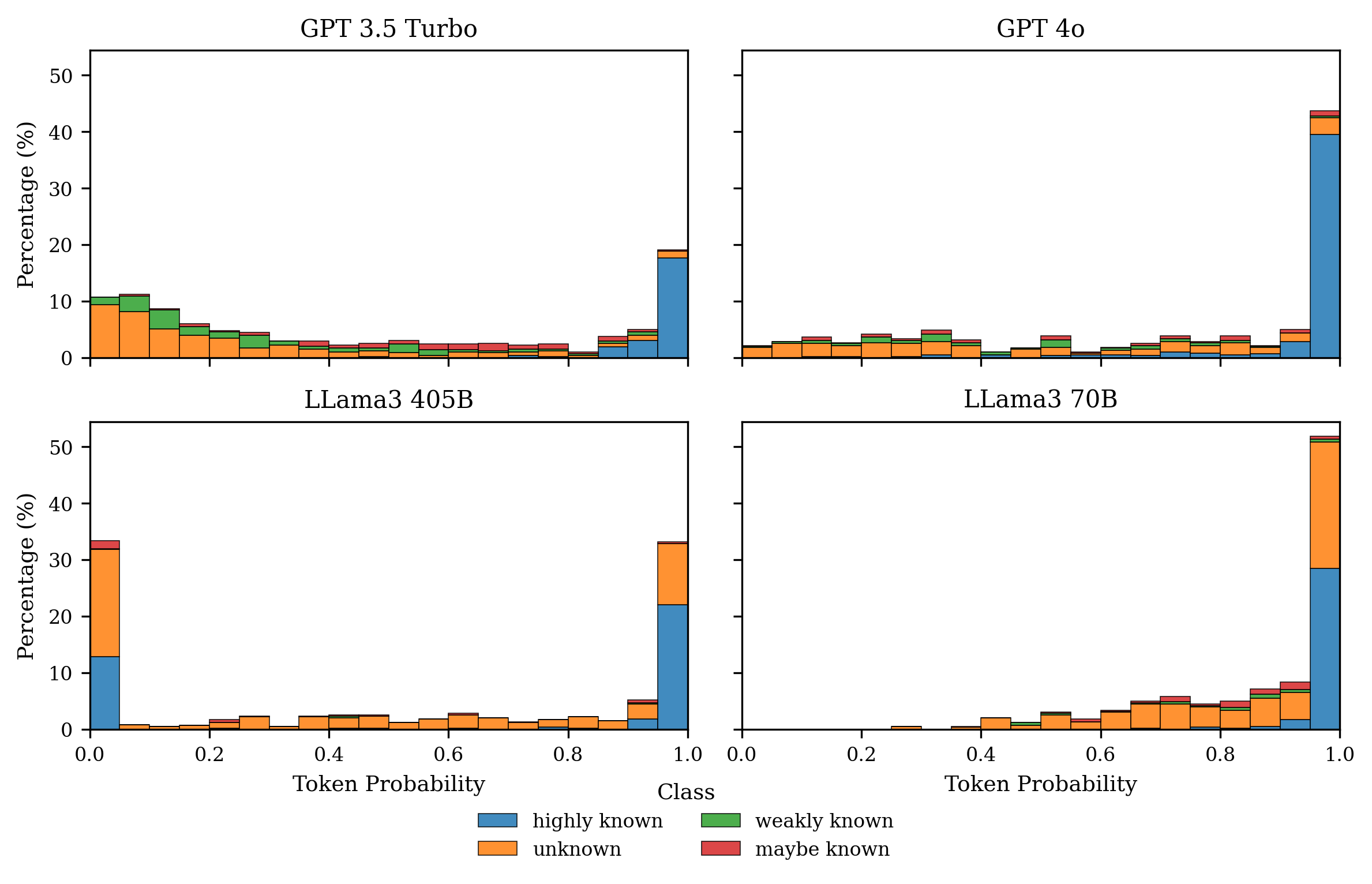}
    \caption{Distribution of self-confidence scores on more models}
    \label{fig:self confidence all}
\end{figure*}

\section{Limitations and Future Work}
\label{sec:limitations}
\paragraph{Limitations.} SR-DCR relies on fixed confidence thresholds, which may not generalize optimally across all domains or tasks. Additionally, the current implementation assumes access to deterministic judge behavior, which may not hold in real-time deployment with stochastic models. The asymmetric debate structure also presumes the availability of prior-free inference, which can be challenging in closed-source or limited-access APIs.

\paragraph{Future Work.} Future directions include learning adaptive confidence thresholds, training a dedicated judge model with supervised debate data, and extending SR-DCR to multi-hop reasoning and document-level tasks. Incorporating human-in-the-loop feedback or interactive oversight could further enhance its applicability to high-stakes or ambiguous domains.

\end{document}